# A Gamma-Poisson Mixture Topic Model for Short Text

Jocelyn Mazarura[1], Alta de Waal[12] and Pieter de Villiers[3]

[1]Department of Statistics, University of Pretoria, Pretoria, South Africa.

[2]Center for Artificial Intelligence Research (CAIR), Pretoria, South Africa.

[3]Department of Electrical, Electronic and Computer Engineering, University of Pretoria, Pretoria, South Africa.

Correspondence should be addressed to Jocelyn Mazarura; Jocelyn.Mazarura@up.ac.za

## Abstract

Most topic models are constructed under the assumption that documents follow a multinomial distribution. The Poisson distribution is an alternative distribution to describe the probability of count data. For topic modelling, the Poisson distribution describes the number of occurrences of a word in documents of fixed length. The Poisson distribution has been successfully applied in text classification, but its application to topic modelling is not well documented, specifically in the context of a generative probabilistic model. Furthermore, the few Poisson topic models in literature are admixture models, making the assumption that a document is generated from a mixture of topics. In this study, we focus on short text. Many studies have shown that the simpler assumption of a mixture model fits short text better. With mixture models, as opposed to admixture models, the generative assumption is that a document is generated from a single topic. One topic model, which makes this one-topic-per-document assumption, is the Dirichlet-multinomial mixture model. The main contributions of this work are a new Gamma-Poisson mixture model, as well as a collapsed Gibbs sampler for the model. The benefit of the collapsed Gibbs sampler derivation is that the model is able to automatically select the number of topics contained in the corpus. The results show that the Gamma-Poisson mixture model performs better than the Dirichlet-multinomial mixture model at selecting the number of topics in labelled corpora. Furthermore, the Gamma-Poisson mixture produces better topic coherence scores than the Dirichlet-multinomial mixture model, thus making it a viable option for the challenging task of topic modelling of short text.

## Introduction

Topic modelling is a text mining technique used to uncover latent topics in large collections of documents. The Latent Dirichlet allocation (LDA) [1] model is the state-of-the art topic model. It has a proven history of success on long documents, such as news articles and e-books. Owing to the increasing popularity of micro-blogging websites, social media platforms and online shopping (which involves product reviews), text that is significantly shorter has become increasingly relevant. Such sources of text potentially hold valuable information that can be useful in many applications, such as event tracking [2], interest profiling [3] and product recommendation [4].

Traditional topic models infer topics based on word co-occurrence relationships between words [5]. In order to extract meaningful topics, a topic model must successfully infer these relationships from a corpus. Per definition, short text contains few words and consequently tends to contain less co-occurrence information than long text. For instance, some platforms, such as Twitter, impose a character limit on each post, which severely constrains the amount of information one can share in a single post. This has created a need to reconsider topic model assumptions in order to overcome this challenge. One topic model which has shown better performance on short text than LDA is the Dirichlet-multinomial mixture model (DMM) [6], [7]. LDA is sometimes referred to as an admixture model [8] as it assumes each document contains several topics. In contrast, DMM is inherently a mixture model, thus it assumes that each document contains only a single topic, which is a seemingly more sensible assumption for short text. This simple assumption is likely the reason for the better performance which has been observed on some short text datasets [9]–[11].

The conjugate Dirichlet prior allows for convenient hierarchical Bayesian modelling of count data using a multinomial distribution, and, over the years, most topic models have been built under the assumption that documents are sampled from a multinomial distribution. Another natural choice of distribution for count data is the Poisson distribution. However, it has received significantly less attention, as some researchers have found that it does not fit natural text [12]. Nevertheless, other researchers have found that the family of Poisson distributions produces good results on text (and other count data) categorisation, which is the motivation for our investigation into the Poisson distribution as a viable option for topic modelling of short text [13], [14].

The contributions of this work are as follows:

1. We propose a new topic model for short text, the Gamma-Poisson mixture (GPM) topic model, that has not been applied in the literature before. This model is based on the Poisson distribution and we show that it is able to produce topics with improved coherence scores when compared to GSDMM (the collapsed Gibbs sampler version of DMM) [6].
2. We derive a collapsed Gibbs sampler for the estimation of the model parameters. This estimation procedure enables the model to estimate the number of topics automatically.
3. We perform extensive experiments in Python on three short text corpora and report on the characteristics of the new model.
4. We have also made available the development version of the GPM model in a Python package at https://github.com/jrmazarura/GPM.

## Related work

Conventional topic models take advantage of word co-occurrence information in documents to infer the latent topics. However, due to its length, this kind of information is limited in short text, which poses a challenge when applying traditional topic models. It is for this reason that short texts are often described as being sparse. In order to overcome the challenges associated with topic modelling of short text, some researchers have proposed pooling or aggregating short texts to create longer documents prior to applying traditional topic models [3], [15]–[18]. Others have successfully proposed modifications to conventional topic models, such as LDA or DMM. These modifications include, incorporating auxiliary information from external corpora [19], [20] and inducing sparsity into the models [21]–[23]. Lastly, another popular approach, is the derivation of completely new models [5], [24]. In light of the success of DMM on short text, the new model that we propose is a modification of DMM.[1]

In the context of topic modelling, the multinomial distribution is most commonly used to model the words in a document. In contrast, significantly fewer topic models are constructed based on the Poisson distribution. Yet, in other text mining fields, such as in text classification [13], [14] and information retrieval [27], some researchers were able to obtain improved results with the Poisson distribution in comparison to the multinomial distribution in their applications. This serves as further motivation for considering the Poisson distribution as a basis for our topic model.

The Gamma-Poisson (GaP) model [28] and the Poisson decomposition model [29] are both examples of topic models that assume word counts follow a Poisson distribution. Other Poisson-based topic models are presented in [30], [31]. None of these models was specifically designed for short text and the authors only test their models on long documents. Our model is distinctly different from these in that it assumes each document contains a single topic, whereas these models assume each document contains multiple topics. The Poisson-based Dirichlet multinomial mixture model (PDMM) [11] is another DMM-based topic model formed by incorporating a Poisson distribution in the generative process so as to allow each document to contain either 1, 2 or 3 topics. In order to allow PDMM to also take advantage of semantic relations between words, Li et al. [11] extended PDMM by incorporating word embeddings through a generalized Pólya urn. Despite PDMM being termed "Poisson-based", it still models word counts with a multinomial distribution.

Lastly, unlike the multinomial distribution, the Poisson distribution does not assume that occurrences of the same word are independent of each other [32]. Furthermore, as the Poisson distribution only has a rate parameter, the need to estimate the total number of trials, which is a non-trivial task, is evaded [14]. In light of these properties, we believe that a Poisson-based topic model could yield favourable results. In the next section, we investigate the characteristics of word frequencies in our datasets to further motivate the case for the Poisson distribution.

## Empirical analysis of word occurrences in short text

In contrast to the amount of literature available on multinomial-based topic models, there is significantly less research on topic models that are based on the Poisson distribution. This is likely due to the work of Gale and Church [12] which demonstrated that the Poisson distribution is not a good fit for observed word frequencies in real world texts. They proposed a $K$-Poisson mixture as a more suitable alternative. In order to motivate that the Poisson distribution fails to model word frequency, Gale and Church [12] selected a word from their corpus, "said", and plotted the graph shown in Figure 1.

Figure 1 shows the number of documents in which the word "said" was used 0 times, 1 time, 2 times, …, or 32 times. The curve shows the predicted number of documents from a Poisson distribution calculated using the maximum likelihood estimate of the parameter. It is clear that the Poisson does not provide a good fit, thus they proposed a mixture of Poisson distributions or a negative binomial distribution as a better alternative. However, it must be noted that the documents under consideration were long and different results may be observed when the same graph is plotted for a word in a short text corpus. To demonstrate this, we selected the word "jet" from topic 1 in the Pascal Flickr corpus (discussed in the Dataset section) and obtained the results in Figure 2.

---

[1] The avid reader is referred to the following recent review papers for further reading on short text topic modelling: [25] and [26].

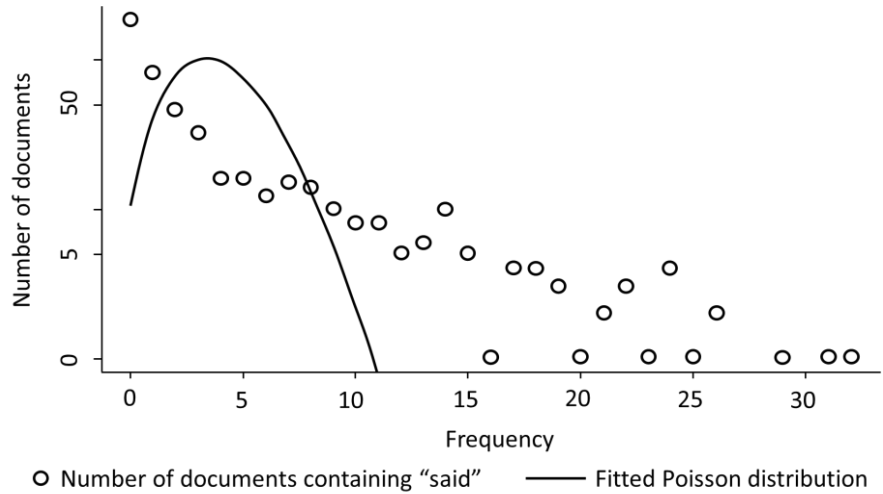

Figure 1: The circles show the number of documents that contain the word "said" for different frequencies. The curve denotes predicted frequencies from a Poisson distribution fitted to the data. Adapted from "Poisson mixtures" by Church, K. W., & Gale, W. A. (1995). Natural Language Engineering, 1(2), 163-190. Copyright by Cambridge University Press 1995. Reproduced with permission.

The length of the documents considered in Figure 1 was approximately 2 000 words per document whereas the average length of a document in the Pascal Flickr corpus was merely 5 words with a minimum and maximum length of 1 and 19 words, respectively. Considering this, it is highly unlikely that large frequencies would be observed. From Figure 2, the maximum frequency of the word "jet" is 1 and, as we no longer have the heavy tail, the predicted values from the Poisson distribution (solid line) are close to the observed values. Similar results were observed with many of the other words in the corpus. Thus, we did not deem it necessary to model each word count with a $K$-Poisson mixture as proposed by Church and Gale [12].

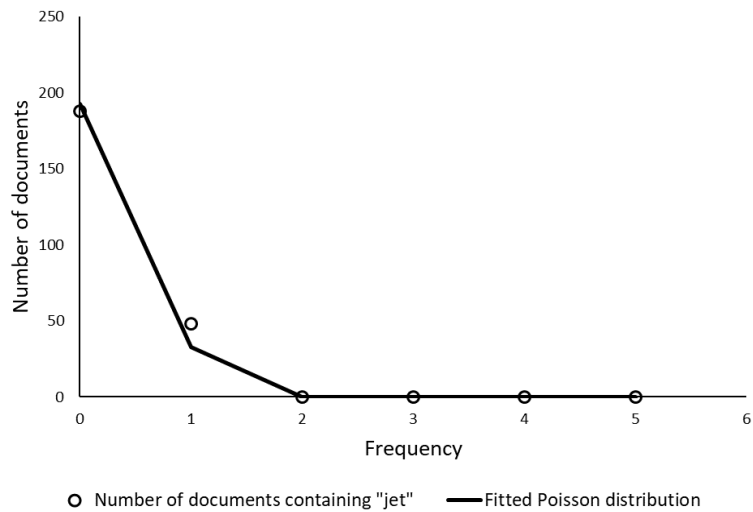

Figure 2: The circles show the number of documents that contain the word "jet" for different frequencies in the Pascal Flickr dataset. The curve denotes predicted frequencies from a Poisson distribution fitted to the data.

Another possible discrete distribution that may be considered is the negative binomial distribution. It is able to relax the assumption made by the Poisson distribution that the mean and variance of the data are equal. The negative binomial is the preferred choice when the observed data displays overdispersion, that is, when the variance exceeds the mean. Further investigation of the means and variances of the words in topic 1 of the Pascal Flickr dataset, as well as several other short text corpora, did not yield significant evidence to warrant the use of the negative binomial distribution.

Gale and Church [12] also identified a phenomenon referred to as burstiness. A word is said to be bursty or contagious if, after its first mention, it is likely to be observed again in the same document. In order to address word burstiness in the context of document classification, some authors have proposed the use of the Dirichlet compound multinomial [33] whereas others suggest using contagious distributions, such as the negative binomial distribution, to model word frequencies [14]. However, Figure 2 does not appear to display evidence of burstiness, neither did we observe evidence of significant burstiness in the short text corpora we studied.

In conclusion, we can see that a simple Poisson is a viable option to model word frequency in short text. In addition, it is also an attractive choice as it has a conjugate prior whereas the $K$-Poisson mixture and negative binomial suggested by Church and Gale [12] do not. It is for these reasons that we consider topic modelling using the Poisson distribution. We now introduce a new topic model, the Gamma-Poisson mixture (GPM) topic model.

## The Gamma-Poisson Mixture Topic Model for Short Text

Table 1 shows a summary of the notation that will be used throughout this paper.

Table 1: Notation.

| Symbol | Description |
|--------|-------------|
| $M$ | number of documents in the corpus. |
| $V$ | size of the vocabulary |
| $K$ | number of topics |
| $N_m$ | length of $m^{\text{th}}$ document ($m = 1, 2, \dots, M$) |
| $\boldsymbol{x}$ | collection of documents |
| $\boldsymbol{x}_m$ | frequency vector of $m^{\text{th}}$ document |
| $x_{mv}$ | number of times word $v$ occurs in the $m^{\text{th}}$ document ($v = 1, 2, \dots, V$) |
| $\boldsymbol{z}$ | vector of topic assignments of each document |
| $z_m$ | topic assignment of document $\boldsymbol{x}_m$ |
| $m_k$ | number of documents in topic $k$ ($k = 1, 2, \dots, K$) |
| $n_{kv}$ | number of times word $v$ is observed in topic $k$ |
| $n_k$ | number of words in topic $k$ |
| $a^{(m)}$ | if $a$ is a quantity that describes a characteristic of the corpus, $a^{(m)}$ denotes the same characteristic of the corpus excluding the $m^{\text{th}}$ document |

The Gamma-Poisson mixture topic model is a hierarchical Bayesian model for topic modelling of short text. For simplicity, it assumes that the frequencies of words in a document are independent of each other and that the corpus is a mixture of documents, which belong to different topics. Mixture models are amongst the simplest of latent variable models.

Considering the success of GSDMM[2] on short text [9]–[11], our GPM topic model makes similar assumptions: (1) Documents are formed from a mixture model and (2) each document belongs to exactly one topic (cluster). This embodies the following probabilistic generative process for a document, $x_m$:

1. A topic, $k$, is randomly selected depending on mixing weights $p(z = k)$.
2. A document is then randomly selected from $p(x_m | z = k)$.

Consequently, the likelihood of a document is given by

$$p(x_m) = \sum_{k=1}^{K} p(x_m | z = k) p(z = k),$$

where $K$ denotes the total number of topics in the corpus. Like GSDMM, GPM makes the Naïve Bayes assumption: given the topic, the frequency of the words in the document are independent of each other. Thus, under GPM the conditional probability of a document given a topic is given by

$$p(x_m | z = k) = \prod_{v=1}^{V} p(x_{mv} | \lambda_{kv}),$$

where $x_{mv}$ denotes the frequency of word $v$ in document $x_m$, and $\lambda_{kv}$ denotes the expected frequency of word $v$ on topic $k$. The key difference between GPM and GSDMM arises at this point. The GPM assumes the frequencies, $x_{mv}$, are modelled according to independent Poisson distributions as opposed to modelling the joint distribution of the counts with a multinomial distribution as in the DMM. In addition, owing to its conjugacy, a gamma prior with shape parameter $\alpha_k$ and scale parameter $\beta_k$ is imposed on $\lambda_{kv}$. Similarly, owing to the Dirichlet distribution's conjugacy to the multinomial, GSDMM assumes a Dirichlet prior.

Under the GPM, the mixing weights represent the proportion of each of the $K$ topics in the corpus. The topic assignment $z$ of each document is modelled by a multinomial distribution. Thus, $p(z = k) = \pi_k$ where $0 \leq \pi_k \leq 1$ and $\sum_{k=1}^{K} \pi_k = 1$. Furthermore, a Dirichlet prior with parameter $\gamma$ is imposed on $\boldsymbol{\pi} = [\pi_1, \pi_2, \ldots, \pi_K]$. As GPM is inherently a mixture model, this part of the model is the same as GSDMM.

The generative process of GPM can be summarised in a graphical model as in Figure 3.

---

[2] GSDMM is the collapsed Gibbs sampler version of DMM [6]. The authors in [6] used the abbreviation GSDMM (Gibbs Sampler DMM). We use this version in this paper, thus from here onwards we will also refer to the GSDMM as opposed to DMM.

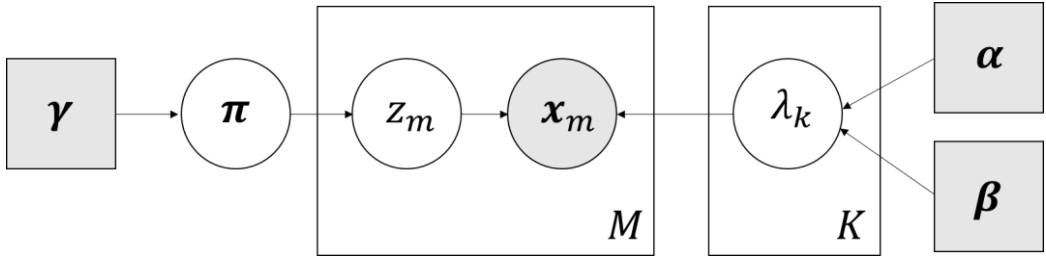

Figure 3: Graphical model of GPM. Shaded squares indicate fixed parameters. Shaded circles denote observed variables, such as a document, $x_m$, and unshaded circles represent latent variables, such as the topic distribution, $\lambda_k$. Rectangles represent repeated structures, whereas arrows indicate conditioning, such as the conditioning of documents on both topic distribution and topic assignment.

Figure 3 describes the statistical conditioning of variables on their parent variables. The only random variable that is observed is the corpus, whereas all others are latent variables. In the following section, we will discuss the estimation procedure for the GPM.

**The Collapsed Gibbs sampler**

A typical Gibbs sampler [34] requires that each parameter be sampled in turn conditioned on all the other parameters. As the topics are only dependent on the topic assignment of each document, it is only necessary to sample the topic assignments. The conjugacy of the chosen priors introduces analytic tractability that makes it is possible to easily integrate out the other parameters that would otherwise need to be sampled. Thus, owing to this simplification, this sampling scheme is referred to as a collapsed Gibbs sampler.

Other estimation techniques, such as the Expectation-Maximisation (EM) algorithm, could have also been used. However, it is the collapsed Gibbs sampler that gives our model the favourable property of being able to automatically select the number of latent topics. In practice, one popular way of selecting the number of topics is achieved via the use of non-parametric topic models [35]. Thus, although parametric in nature, our model displays this "non-parametric" behaviour to some extent.

In order to estimate the model parameters, the collapsed Gibbs sampler assigns each document to a single topic. This is achieved by sampling from the conditional probability of document $x_m$ belonging to a class, $p\big(z_m = z \big| z^{(m)}, x, \alpha, \beta, \gamma\big)$. From the rules of conditional probability, it follows that

$$p\big(z_m = z \big| z^{(m)}, x, \alpha, \beta, \gamma\big) = \frac{p(x, z|\alpha, \beta, \gamma)}{p(x, z^{(m)}|\alpha, \beta, \gamma)} \propto \frac{p(x, z|\alpha, \beta, \gamma)}{p(x^{(m)}, z^{(m)}|\alpha, \beta, \gamma)}, \qquad (1)$$

where the superscript $(m)$ is used to denote that document $x_m$ is excluded. $\alpha = [\alpha_1, \alpha_2, \dots, \alpha_V]$ and $\beta = [\beta_1, \beta_2, \dots, \beta_V]$ are the hyperparameters of the gamma prior, whereas $\gamma = [\gamma_1, \gamma_2, \dots, \gamma_K]$ denotes the hyperparameter of the Dirichlet prior.

In order to sample a topic assignment for each document according to Equation 1, only the joint distribution, $p(x, z|\alpha, \beta, \gamma)$, is required. It is shown in the Appendix that it is given by

$$p(\boldsymbol{x}, \boldsymbol{z} | \boldsymbol{\alpha}, \boldsymbol{\beta}, \boldsymbol{\gamma}) = \frac{\Delta(\boldsymbol{m} + \boldsymbol{\gamma})}{\Delta(\boldsymbol{\gamma})} \prod_{k=1}^{K} \prod_{v=1}^{V} \frac{\Gamma(n_{kv} + \alpha_v)}{\boldsymbol{x}! \, \Gamma(\alpha_v)} \times \frac{\beta_v^{n_{kv}}}{(m_k \beta_v + 1)^{n_{kv} + \alpha_v}}. \tag{2}$$

By substituting Equation 2 into Equation 1, under the assumption that $\alpha_v = \alpha$, $\beta_v = \beta$ and $\gamma_k = \gamma$ for all $v$ and $k$, it follows that Equation 1 can be expressed as

$$p\big(\boldsymbol{z}_m = z | \boldsymbol{z}^{(m)}, \boldsymbol{x}\big)$$

$$\propto \frac{m_z^{(m)} + \gamma}{M - 1 + K\gamma} \times \frac{\beta^{N_m}}{\boldsymbol{x}_m!} \times \frac{\left(m_z^{(m)} \beta + 1\right)^{n_z^{(m)} + V\alpha}}{\left(m_z^{(m)} \beta + \beta + 1\right)^{n_z^{(m)} + N_m + V\alpha}} \times \prod_{v=1}^{V} \prod_{j=1}^{x_{mv}} \Big(n_{zv}^{(m)} + \alpha + j - 1\Big). \tag{3}$$

Thus, for each document, a topic is sampled repeatedly until convergence is achieved. The topics are then found by the following estimates:

$$\hat{\lambda}_{kv} = \frac{n_{kv} + \alpha_v}{\left(m_k + \dfrac{1}{\beta_v}\right)},$$

where the top words that describe topic $k$ are the words with the highest estimated expected frequencies, $\hat{\lambda}_{kv}$. Full details of the derivation are given in the Appendix.

The collapsed Gibbs sampler for GPM is summarised in Algorithm 1.

---

Algorithm 1: Collapsed Gibbs sampler for GPM.

**Input**: Corpus, $\boldsymbol{x}$
**Output**: Topic labels for each document, $\boldsymbol{z}$
**Begin**

> Initialise $m_z, n_z$ and $n_{zv}$ to 0 for each topic $z$
> **for** each document $\boldsymbol{x}_m, m = 1, 2, \ldots, M$
>> randomly sample a topic for $\boldsymbol{x}_m$
>> $z_m \leftarrow z \sim Categorical(1/K)$
>> $m_z \leftarrow m_z + 1$ and $n_z \leftarrow n_z + N_m$
>> **for** each word frequency $x_{mv}$ in $\boldsymbol{x}_m$
>>> $n_{zv} \leftarrow n_{zv} + x_{mv}$
>
> **for** $i = 1, 2, \ldots, I$ iterations
>> **for** each document $\boldsymbol{x}_m, m = 1, 2, \ldots, M$
>>> record the current topic of document $\boldsymbol{x}_m$: $z = z_m$
>>> $m_z \leftarrow m_z - 1$ and $n_z \leftarrow n_z - N_m$
>>> **for** each word frequency $x_{mv}$ in $\boldsymbol{x}_m$
>>>> $n_{zv} \leftarrow n_{zv} - x_{mv}$
>>>
>>> sample a new topic for $\boldsymbol{x}_m$
>>> $z_m \leftarrow z \sim p\big(\boldsymbol{z}_m = z | \boldsymbol{z}^{(m)}, \boldsymbol{x}\big)$ (Equation 3)
>>> $m_z \leftarrow m_z + 1$ and $n_z \leftarrow n_z + N_m$
>>> **for** each word frequency $x_{mv}$ in $\boldsymbol{x}_m$
>>>> $n_{zv} \leftarrow n_{zv} + x_{mv}$

---

Topic models are very powerful tools as they possess characteristics from both clustering and dimensionality reduction techniques: (1) A corpus is represented in a lower dimensional form by a set of topics and, (2) similar to clustering, each document is associated with a single topic or multiple topics depending on the model. Our GPM topic model possesses both these qualities. The first property is captured by the $\lambda$ parameters. The second is satisfied in Equation 3. The advantage of topic models over traditional clustering algorithms, is that "labels" are also produced, in the form of topics. In order for topic models to be useful, they are designed to not only provide data compression, but to also produce interpretable topics.

In order to demonstrate the utility of our new model, we perform extensive experimentation on different datasets. Details are provided in the next section.

## Experiments

### Datasets

In order to test our model, we ran experiments on different datasets and compared the performance of GPM against that of GSDMM. The datasets have been summarised in Table 2. All statistics were collected from the datasets after basic pre-processing (removal of stop words, punctuation, special symbols and numbers).

The Tweet dataset [6] is a collection of tweets from the 2011 and 2012 Text REtrieval Conference. The most relevant tweets in 89 different categories were selected to create this collection. Each tweet is regarded as an individual document.

The Pascal Flickr dataset contains captions of images from Flickr and the Pattern Analysis, Statistical Modelling, and Computational Learning (PASCAL) Visual Object Classes Challenge [36]. The captions are divided into 20 different classes and altogether the corpus contains 4 821 captions which are each treated as individual documents.

The Search Snippet dataset [37] was created by first selecting 8 different domains: Business, Computers, Culture-Arts-Entertainment, Education-Science, Engineering, Health, Politics-Society and Sports. For each domain, 11 to 118 related phrases were typed into the Google search engine, and then the snippets from the top 20 to 30 results were collected to create a corpus of 12 295 snippets.

Note, we will often refer to the original number of classes/categories for each dataset as the true number of topics/clusters or true $K$.

Table 2: Document statistics.

| Dataset | Number of documents | Size of vocabulary | Number of topics | Average (Standard deviation) of document length | Minimum (Maximum) document length |
|---|---|---|---|---|---|
| Tweet | 2 472 | 5098 | 89 | 8.5 (3.2) | 2 (20) |
| Pascal Flickr | 4 821 | 3188 | 20 | 4.9 (1.8) | 1 (19) |
| Search Snippets | 12 295 | 4705 | 8 | 14.4 (4.4) | 1 (37) |

All datasets can be obtained from https://github.com/qiang2100/STTM [25].

**Experimental design**

All experiments were executed in Python 3.6 in Windows 10 on a computer with a 3.50 GHz quad core processor and 16 GB RAM. We used our own implementations of each model and have made our implementation of the GPM topic model publicly available as a Python package at https://github.com/jrmazarura/GPM. For the GSDMM, the parameter values were set to $\alpha = \beta = 0.1$ and the algorithm was run for 15 iterations, as in the original paper. For the GPM, the $\gamma$ parameter plays the same role as the $\alpha$ parameter in GSDMM, thus it was also set to 0.1. For the Poisson distribution we opted for a gamma prior with shape and scale parameters, $\alpha$ and $\beta$, both set to 0.001. This choice is motivated within the upcoming sections.

**Document Length Normalisation**

Since the Poisson distribution gives the probability of observing a given number of events in a fixed interval, it is necessary to normalise the lengths of the documents. This is achieved by replacing the word frequencies, $x_{mv}$, with

$$x_{mv}^{\text{new}} = \frac{N x_{mv}}{\sum_{v=1}^{V} x_{mv}},$$

where $N$ denotes a predefined length [13]. In all our experiments, we set $N = 20$ and rounded off each $x_{mv}^{\text{new}}$ to the nearest integer. $N$ was chosen to be 20 as it provided a good balance between performance and runtime for our datasets.

**Model Evaluation**

In order to evaluate the performance of our model we used the average of the UMASS topic coherence [38] score for each topic. The coherence score for each topic, $T$, is given by

$$coherence(T) = \sum_{(v_i, v_j) \in T} \log \frac{D(v_i, v_j) + \epsilon}{D(v_j)},$$

where $v_i$ denotes the $i^{\text{th}}$ word in topic $T$, $D(v_i, v_j)$ denotes the number of documents in which words $v_i$ and $v_j$ co-occur and $D(v_j)$ denotes the number of documents in which word $v_j$ occurs. $\epsilon$ is a smoothing parameter to prevent taking the logarithm of zero and it is set to equal 1 as proposed in the original paper. As with most topic models, the GPM is an unsupervised technique. Model evaluation is generally not a trivial task in the context of unsupervised learning as datasets lack labels upon which evaluations can be based. The UMASS coherence score is a well-known measure of the degree of interpretability of a topic and it has been shown to align well with human evaluations of coherence [38]. Naturally, topics that are coherent are most desirable; therefore, a higher average coherence score is preferable. Similar to GSDMM, our model has the special characteristic of being able to automatically select the number of topics, thus, the coherence score is only calculated on the topics found by the model.

## Results and Discussion

### Influence of the starting number of topics

Topic modelling is typically an unsupervised technique. Similar to $K$-means clustering, the number of topics (clusters), $K$, is a challenge to select as the value is not usually known in advance. The GPM is able to infer the number of topics automatically provided that the starting value of $K$ is large enough. This is due to the dependence of the topic assignment probability, Equation 3, on $m_k$, which is the number of documents in topic $k$. This implies that a document is more likely to be assigned to a topic which has documents assigned to it, than a topic that does not have documents assigned to it.

As will be shown in the next section, the collapsed Gibbs sampler is quick to converge, thus the Gibbs sampler was run for 15 iterations. As our model also provides stable and relatively consistent results (as will be shown in the next section), experiments were repeated 3 times assuming $K = 5, 10, 20, 30, 40, 50, 100, 200, 300, \ldots, 800$. We set $\alpha_v = \beta_v = 0.001$ for all $v$ and $\gamma_k = 0.1$ for all $k$. Table 3 shows the average number of topics found by the model for some of the different starting values of $K$, whereas Table 4 shows the corresponding average coherence scores.

Table 3: Average final number of topics found by the model (and standard deviation).

| Dataset | True $K$ | Starting value of $K$ | | | | |
|---|---|---|---|---|---|---|
| | | 50 | 100 | 200 | 400 | 800 |
| Tweet | 89 | 42(2) | 61 (3) | 67 (7) | 76 (5) | 77 (5) |
| Pascal Flickr | 20 | 14 (2) | 26 (2) | 33 (6) | 33 (5) | 39 (3) |
| Search Snippets | 8 | 19 (6) | 16 (4) | 25 (5) | 26 (3) | 32 (1) |

Table 4: Average topic coherence scores (and standard deviation).

| Dataset | Starting value of $K$ | | | | |
|---|---|---|---|---|---|
| | 50 | 100 | 200 | 400 | 800 |
| Tweet | -25.02 | -23.76 | -19.92 | -18.89 | -18.13 |
| | (0.91) | (1.65) | (1.47) | (0.06) | (0.58) |
| Pascal Flickr | -37.01 | -34.50 | -33.85 | -30.39 | -31.53 |
| | (4.16) | (2.06) | (2.92) | (2.59) | (1.78) |
| Search Snippets | -50.56 | -50.17 | -50.71 | -49.47 | -51.19 |
| | (1.85) | (2.06) | (2.26) | (4.15) | (1.81) |

Figures 4 to 6 provide a visual summary of these results. According to Figures 4(a), 5(a) and 6(a), in all cases, the model approaches the true number of topics as the starting number of topics increases. In most cases, the most accurate number of topics was found by setting $K$ to 400. For the Tweet dataset, the model converges to between 70 and 80 topics, which is close to the true value of 89. For the other datasets, the model slightly over-estimates the number of topics. On the Pascal Flickr dataset, at $K = 400$, the final number of clusters is over-estimated by about 10 topics (true $K = 20$) whereas on the Search Snippets dataset, the final number of clusters is over-estimated by about 20 topics (true $K = 8$). One possible reason for this difference could be that the human labelling may have been too rigid and documents were classified into too few topics yet there may have been subtopics present. Consequently, it is possible that such a discrepancy could also arise if different human reviewers were tasked with labelling each document independently. In the context of topic modelling, this difference is not usually a problem, especially if the topics are interpretable, as the model may have simply identified subtopics present in the corpus. Since the model does not differentiate between "main" topics or subtopics, they would all be included together in the final topic count. Nonetheless, it is still striking that in both cases, the model was able to automatically discard the extra 80-90% of topics that were unnecessary. This greatly alleviates the challenge of selecting the appropriate value of $K$.

In topic modelling, one of the most important aspects is the interpretability of the uncovered topics. Even if the final number of clusters found is not necessarily the same as what human annotators would find, it is important that the words in the topics are coherent. Figures 4(b), 5(b) and 6(b) show that the coherence improves as the initial $K$ increases. In fact, there reaches a point where there is almost no more improvement in average coherence when the initial number of topics is increased. In most cases, there appears to be an insignificant improvement to the coherence score, when $K$ is selected to be greater than 200.

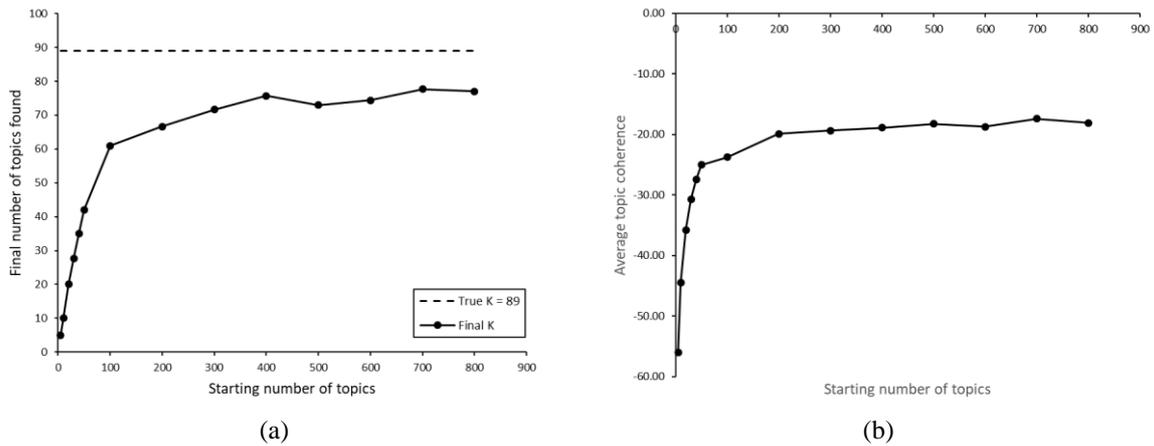

(a)                                                    (b)

Figure 4: Tweet dataset: (a) Average final number of topics found by the model (b) Average topic coherence scores

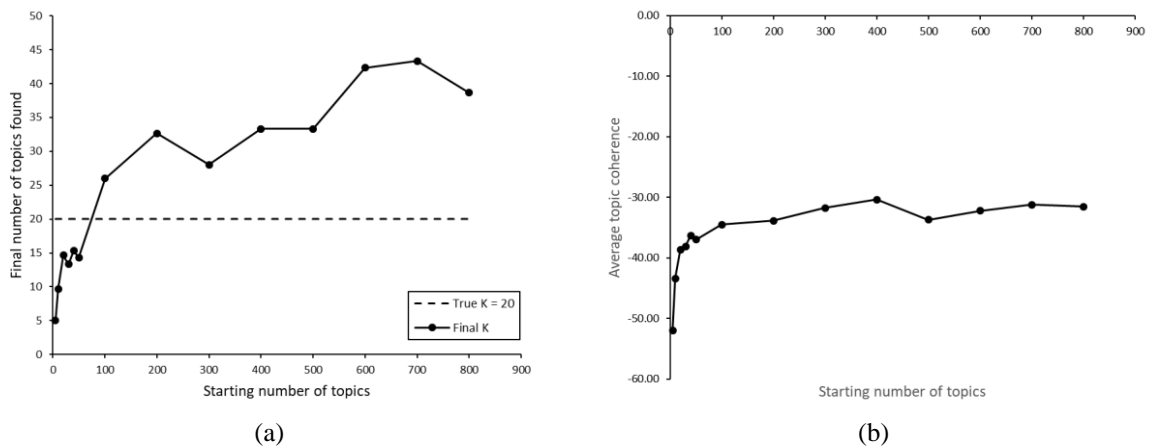

(a)                                                    (b)

Figure 5: Pascal Flickr dataset: (a) Average final number of topics found by the model (b) Average topic coherence scores

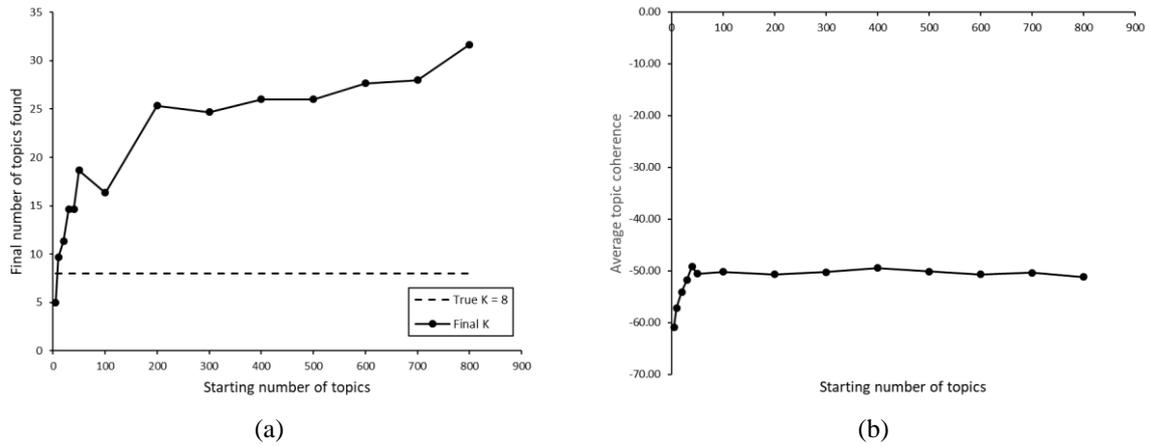

(a) (b)

Figure 6: Search Snippets dataset: (a) Average final number of topics found by the model (b) Average topic coherence scores

## Influence of the number of iterations

One of the challenges faced when using sampling methods to estimate parameters is determining the appropriate number of sampling iterations to perform. In order to investigate the performance of the models with respect to the number of iterations, we recorded the average coherence and number of clusters at each of 30 iterations. This was repeated three times for each dataset. From the previous results, we found that the number of clusters was close to the human annotated number and the coherence scores reached their maximum when the model started with 400 topics, thus we use this value in all the experiments. In addition, we also set $\alpha_v = \beta_v = 0.001$ for all $v$ and $\gamma_k = 0.1$ for all $k$. The results are shown in Figures 7 to 9. The (a) graphs all show the number of clusters that the model found at each iteration, whereas the (b) graphs show the topic coherence at each iteration. In general, similar patterns are observed. It is evident that convergence is reached quickly. In all cases, convergence is reached by the 15th iteration and the variation in the results is typically relatively small.

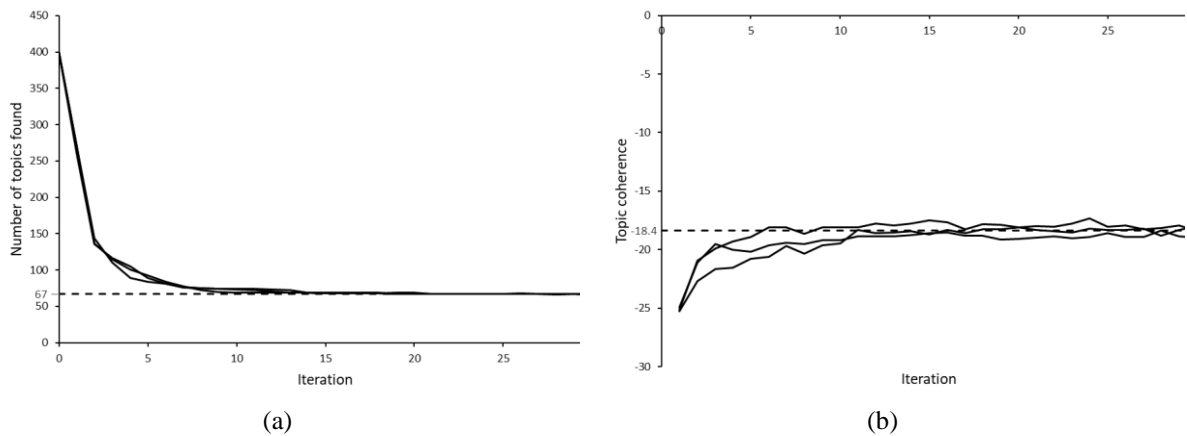

(a) (b)

Figure 7: Tweet dataset: (a) Number of topics found by the model per iteration (b) Average topic coherence score per iteration

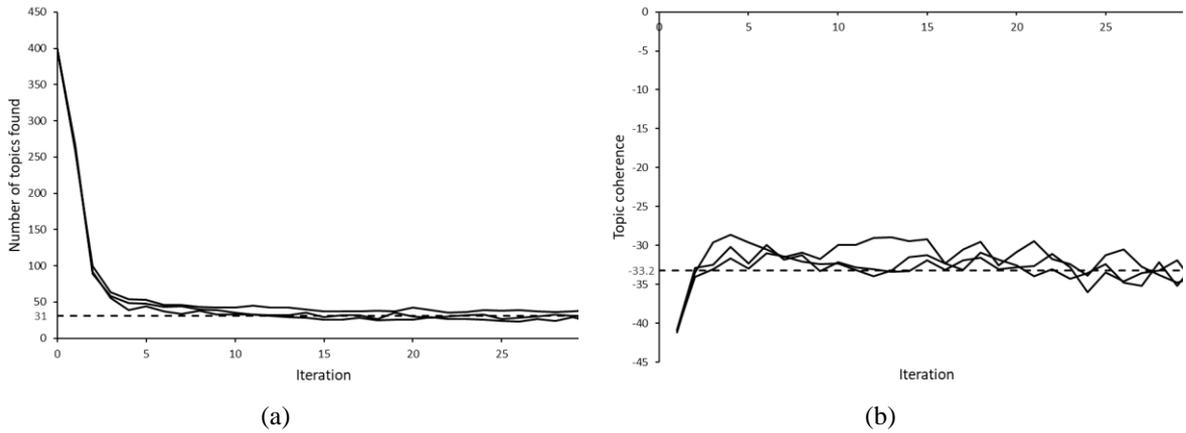

Figure 8: Pascal Flickr dataset: (a) Number of topics found by the model per iteration (b) Average topic coherence score per iteration

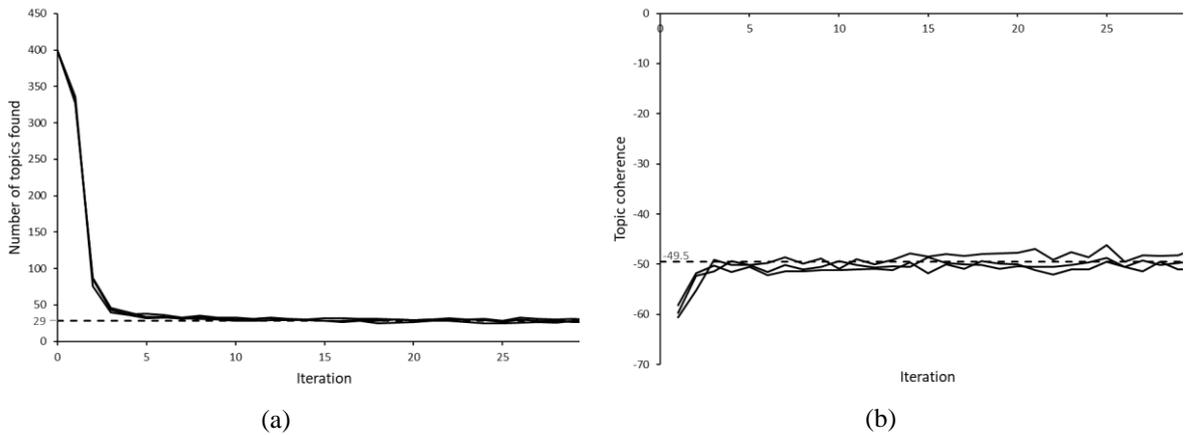

Figure 9: Search Snippets dataset: (a) Number of topics found by the model per iteration (b) Average topic coherence score per iteration

**Influence of alpha and beta**

The hyperparameters $\alpha_v$ and $\beta_v$ represent the shape and scale parameters of the gamma distribution respectively, and $\gamma_k$ represents the hyperparameter of the Dirichlet prior. We assume that $\alpha_v = \alpha, \beta_v = \beta$ and $\gamma_k = \gamma$ for all $v$ and $k$. The $\gamma$ parameter is analogous to the $\alpha$ parameter of the GSDMM. The authors of GSDMM conducted experiments to investigate the impact of different selections of $\alpha$ on the number of clusters found and they observed that it had a very small impact. As expected, we also observed similar results with GPM thus we only focus on the impact of $\alpha$ and $\beta$, assuming $\gamma = 0.1$. The GPM was run on the Pascal Flickr dataset for $K = 40, \alpha = 0.01, 0.05, 0.25, 0.5, 0.75, 1, 2$ and $\beta = 5, 2, 1, 0.5, 0.2$. Then the final number of clusters found was recorded. The results on the Pascal Flickr dataset are shown in Figure 10.

Owing to the computationally heavy nature of performing a grid search, each experiment was run only once per pair of $\alpha$ and $\beta$ values, with the starting number of topics set to be at least 20 more than the true value. Figure 10 shows a clear downward trend for all values of $\beta$, the scale parameter. However, the final number of topics found is clearly influenced by the shape parameter, $\alpha$. On the Pascal Flickr dataset, the model was only able to get close to the true number of topics (20) when $\alpha$ was chosen to be near 0.5. Similar downward trends were also observed on the other two datasets and $\beta$ was also found to be of minimal impact on the number of topics found. However, for the Tweet dataset, $\alpha$ was required to be near 0.05 for the model to find close to 89 topics, whereas the Search Snippets dataset required an $\alpha$ value close to 1.5 to find close to 8 topics. Figure 11 shows the probability density functions of the gamma distributions with these different values of $\alpha$ and a fixed value of $\beta = 0.5$. It is evident that these choices of alpha tend to produce skewed distributions, which place most of their probability near 0.

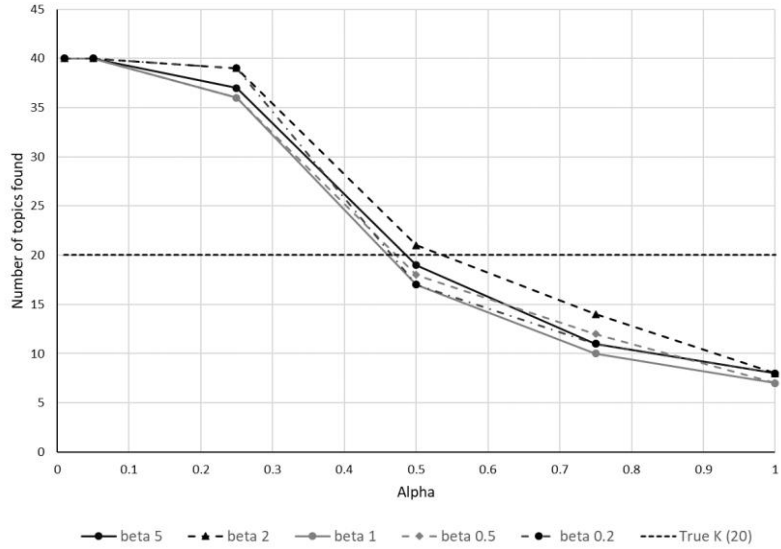

Figure 10: Final number of topics found for different values of alpha and beta on the Pascal Flickr dataset

Based on the chosen values of $\alpha$ and $\beta$, the expected value of the gamma priors for the Tweet, Pascal Flickr and Search Snippets datasets are 0.025, 0.25 and 0.75, respectively. Considering the short length of the documents and the massive sizes of the vocabularies, it is not surprising that most words will have very low observed frequencies. In fact, since many zeros are observed for each word, the estimates of the Poisson parameters are also very small which results in most of the probability being loaded on zero. For example, $p(x) = 0.975$ for $x = 0$ where $X \sim Poi(0.025)$.

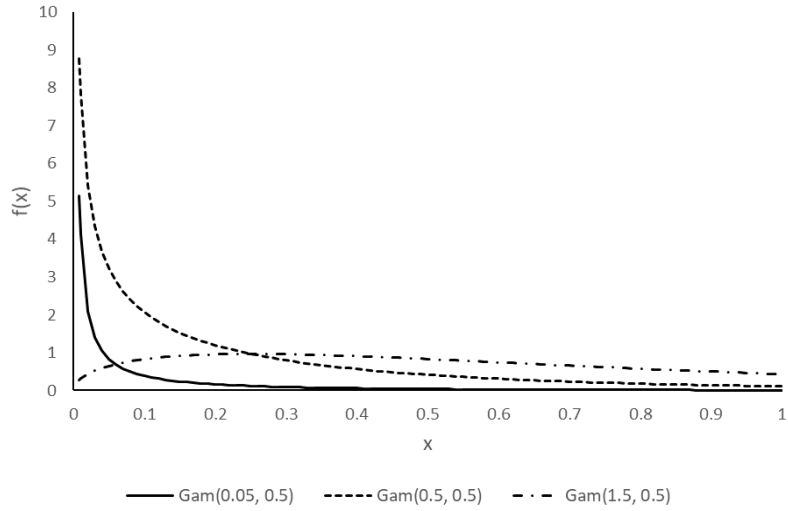

Figure 11: Probability density functions of the gamma distributions for $\alpha = 0.05, 0.5, 1.5$ and a fixed value of $\beta = 0.5$

A similar comparison to that of Figure 10 was also conducted to investigate the impact of $\alpha$ and $\beta$ on the coherence scores and the results from the Pascal Flickr dataset are shown in Figure 12. Figure 12 shows the average coherence of topics found for the same values of $\alpha$, $\beta$ and $K$ used in Figure 10. The labels at each point indicate the number of topics found by the model.

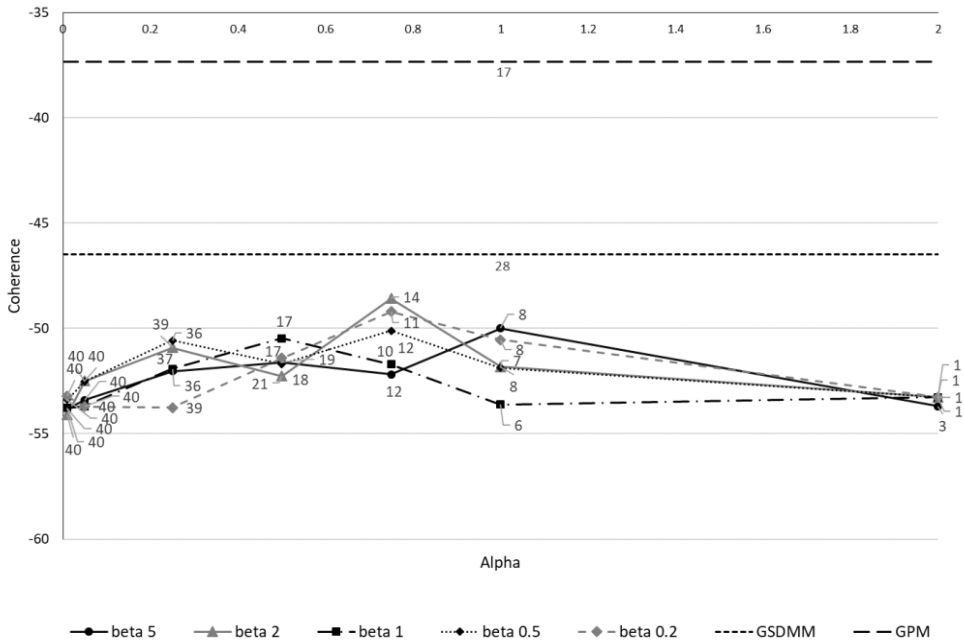

Figure 12: Average topic coherence of topics found for different values of alpha and beta on the Pascal Flickr dataset. The labels at each point indicate the number of topics found by the model.

In Figure 12, a general pattern is observed. The coherence scores appear to increase, then drop as $\alpha$ increases and, once again, $\beta$ does not appear to have a significant impact. The Tweet and Search Snippets datasets also displayed general patterns, but the pattern was not necessarily the same across the datasets. This simply serves as an indication that the selection of $\alpha$ is not a trivial task.

It is evident from Figure 12 that increasing $\alpha$ decreases the number of topics the model tends to find. Conversely, as $\alpha$ decreases, the number of topics found increases. Interestingly, this behaviour of $\alpha$ in GPM is similar to the behaviour that was observed with the $\beta$ parameter of GSDMM [6]. According to Equation 3, for small values of $\alpha$, the probability of a document belonging to a topic is more sensitive to $n_{kv}$, the number of times word $v$ is observed in topic $k$. This means that, when a topic has more words in common with a document it is more likely to be assigned to that topic. On the other hand, when $\alpha$ is large, the probability of being assigned to a topic is less sensitive to $n_{kv}$. Instead, the probability is influenced more by the first term in Equation 3, which is dependent on $m_k$, the number of documents in topic $k$. As a result, a topic with more documents is likely to get larger since Equation 3 will assign more probability to topics that contain more documents. This explains the tendency of the model to assign all the documents to one topic when $\alpha$ is large.

In practice, the number of clusters is not usually known in advance so it is not possible to use the true $K$ to choose a suitable value for $\alpha$. Furthermore, the coherence is also not always highest at the true number of clusters. In order to overcome this challenge, we then considered setting the hyperparameters of the gamma prior to $\alpha = \beta = 0.001$.[3] The top horizontal line in Figure 12 shows the coherence score found by the GPM under this gamma(0.001, 0.001) prior.[4] The coherence is not only higher than that of the other selections of $\alpha$ and $\beta$, but the GPM also outperforms the GSDMM model (indicated by the lower horizontal line). In addition, the average number of clusters found by the GPM was also close to the true value. It is for this reason, that we recommend the use of $\alpha = \beta = 0.001$ and use these values in all our experiments.

In conclusion, it is clear that this selection of $\alpha$ and $\beta$ greatly simplifies the topic modelling process for the GPM. In addition, we have also seen that the model possesses the flexibility of allowing the user to easily adjust the number of topics found by simply changing the value of $\alpha$.

**Comparison with Dirichlet-Multinomial mixture model**

The GSDMM model was originally presented as a clustering algorithm, as opposed to a topic model, and was consequently assessed on its ability to cluster documents [6]. As the GPM is designed for topic modelling, we assess its ability to extract meaningful topics by investigating the topic coherence. Despite there being other topic models for short text, the GPM is related to the GSDMM in that it also makes the one-topic-per-document assumption and is able to automatically select the number of topics. In order to compare the performance of the GPM topic model against that of the GSDMM, both models were fitted to the datasets and the results are summarised in the figures and tables that follow. All models were run for 15 iterations starting with 400 initial topics. This was repeated 10 times for each model with $\alpha_v = \beta_v = 0.001$ for all $v$ and $\gamma_k = 0.1$ for all $k$.

---

[3] In Bayesian literature, the gamma distribution with shape and rate parameters both equal to 0.001 is a commonly used non-informative prior [39]. In this paper, the gamma distribution is parameterised by shape and scale parameters. Despite using the scale-parameter instead of rate-parameter formulation, we show empirically that choosing 0.001 yields better performance than other choices. This is likely because the data contains many zeros and this gamma places most of its probability around 0.

[4] This result is for a fixed alpha and beta, but it is shown in the graph as a horizontal line across all alpha values to emphasize that this choice of parameter outperforms the GPM with other choices of alpha and beta. For ease of comparison, the GSDMM is also indicated by a horizontal line although its hyperparameters are also fixed.

Figure 13 shows boxplots of the topic coherence scores. It is evident that the GPM generally outperforms the GSDMM in all three datasets, as the topic coherence of the topics obtained by the GPM is mostly larger than that of the GSDMM.

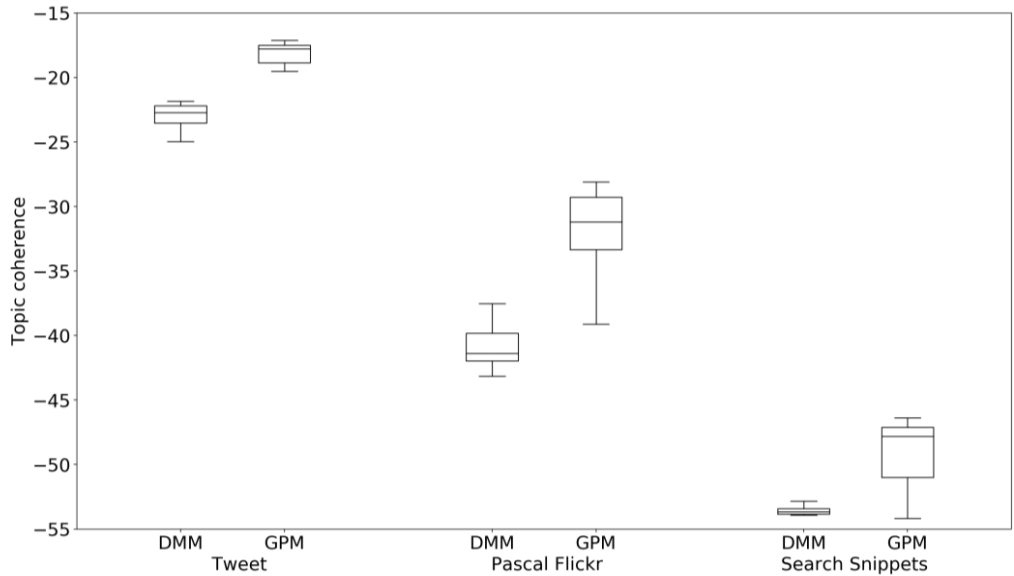

Figure 13: Coherence scores of the different models

For completeness, we also consider the number of clusters found by each model in Table 5. For the Tweet corpus, the true number of clusters, as determined by human annotators, is 89. On average, the GSDMM was more inclined to find more clusters than the GPM. It is also worthwhile to note that the results obtained for the GSDMM on the Tweet dataset are close to those obtained in the original paper [6]. On the Pascal Flickr and Search Snippets datasets, both models tended to find more clusters than those determined by the human annotators. However, the GPM was able to get closer to the true $K$ value than the GSDMM. Interestingly, on the Search Snippets corpus, the GSDMM found significantly more topics than were found by the GPM. It is likely the case that the GSDMM found finer-grained topics, thus increasing the number of topics found, whereas the GPM model discovered fewer, but more general, topics.

Table 5: Summary of number of topics found by each model

| | | GSDMM | | GPM | |
| --- | --- | --- | --- | --- | --- |
| Dataset | True $K$ | Average | Standard deviation | Average | Standard deviation |
| Tweet | 89 | 98 | 3.56 | 75 | 4.42 |
| Pascal Flickr | 20 | 48 | 4.58 | 35 | 5.68 |
| Search Snippets | 8 | 303 | 7.19 | 26 | 2.16 |

We now consider the actual topics found by the models on one of the datasets. We specifically focus on the Search Snippets results in order to observe what other topics were found by the GSDMM model that were not found by the GPM. Table 6 lists some of the top words for each of the topics found by the GPM (column 2), as well as possible labels for each topic (column 1). The labels have been assigned based on the original 8 topics of the dataset and then a possible subtopic label was added in parentheses. This labelling and selection of subtopics was performed subjectively, so another annotator's assessment may produce different results.

Table 6: Topics found by GPM

| Topic (subtopic) | Top words |
| --- | --- |
| Business (software) | trillian instant pro studios creators messenger accounting |
| Business (trade) | import trade export leads business international global |
| Business (consumer) | consumption consumer motives goals ratneshwar glen mick |
| CAE (Chris Pirillio) | pirillo chris live internet broadcast podcast itunes streaming |
| CAE (music) | lyrics song com archive searchable songs database search |
| CAE (painting) | surreal leonardo del vinci picasso surrealism artlex artchive |
| CAE (videos) | videos metacafe ping pong movies internet tags amazing clips |
| CAE (movies) | imdb movies celebs title name diesel movie mtv aesthetic weapon |
| CAE (posters) | posters allposters com prints custom professional framing |
| CAE (transformers) | transformers movie world bay war alien directed races |
| Computers (networking) | approach computer networking featuring ross kurose |
| Computers (root) | root roottalk expression formula cern draw retrieve rene value |
| Computers (programming) | computer programming software web memory wikipedia intel |
| Computers (code) | formula expression kspread value user symbol log api input |
| Computers (connections) | speed test com accurate flash cable speedtest dsl connections |
| ES (news) | information com news wikipedia research edu home science |
| ES (history) | eawc edu classic ancient exploration greece evansville anthony |
| ES (dictionary) | dictionary online definition word christ merriam webster |
| Health (diet) | calorie calories energy drink enviga counter nutrition picnics |
| Health (disease) | treatment arthritis cause symptoms diagnosis lupus disease |
| PS (society) | bombs smoke homepage police press blogspot accounting bank |
| PS (politics) | party bob led revolutionary worker communist revolution |
| Sports (cars) | wheels rims car custom chrome tires truck inch tire |
| Sports (tennis) | match hits russia anna chakvetadze sania financial india |
| Sports (quad biking) | quad china atv automatic reverse quads gear product showroom |
| Sports/Business | goalkeepers cricket nasdaq information stock market security |
| CAE/Computer | span painting election contractors staining servicemagic |

\* Key: CAE = Culture-Arts-Entertainment, ES = Education-Science, PS = Politics-Society

In assigning the topics to the predefined labels, one challenge faced was that some topics had potential overlaps. For instance, a topic in the Engineering category could also have fallen in the Education-Science category. By analysing the first column, we also observe that 7 out of the 8 original predefined topics appear to be represented in these results. According to our labelling, the missing topic is the Engineering topic. This is most likely due to the fact that only 369 of the 12 295 documents belonged to this topic, which is merely 3% of the entire corpus. The proportions of each topic in the Search Snippets corpus are shown in Figure 14.

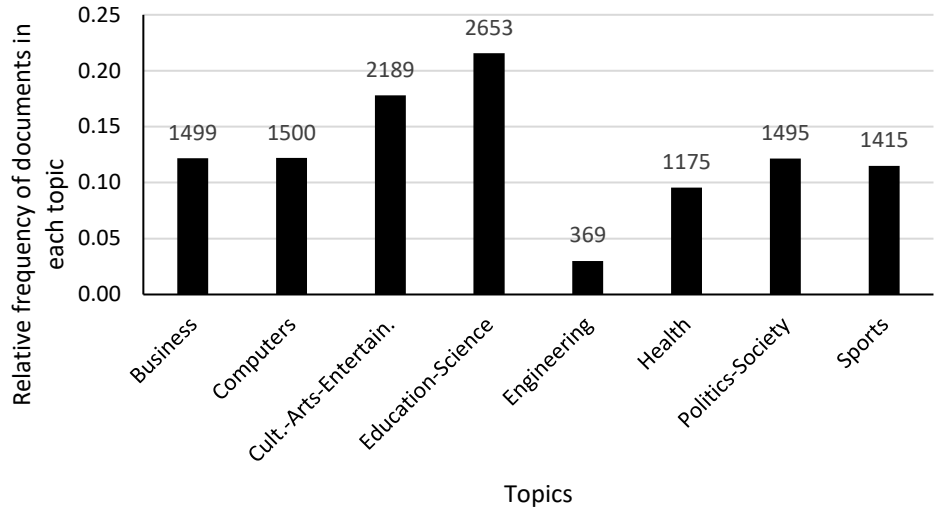

Figure 14: Relative frequency of documents belonging to each topic in the Search Snippets corpus. The number above each bar is the frequency of documents belonging to each topic. The corpus contains a total of 12 295 documents.

As was observed in Table 5, the GSDMM found more than 250 extra topics. Table 7 shows two additional subtopics for each of the 8 predefined categories that were found by the GSDMM, but not the GPM.

Table 7: Selected topics found by GSDMM

| Topic | Top words |
| --- | --- |
| Business (economics) | gdp economy product domestic gross economic value market |
| Business (jobs) | jobs job com search careerbuilder accounting marketing sales sites |
| CAE (fashion) | fashion designers design designer clothing accessories milan clothes |
| CAE (famous places) | ballet hollywood california angeles los universal florida studios |
| Computers (systems) | systems theory analysis design information programming amazon |
| Computers (security) | security computer network spam virus spyware viruses networking |
| ES (genetics) | research national gov laboratory genetic home institute genome |
| ES (earth) | earth structure interior edu crust tectonics model kids gov core |
| Engineering (physics) | physics quantum theory theoretical solid edu research technology |
| Engineering (Einstein) | einstein albert physics nobel eric literature weisstein world time |
| Health (aids) | hiv aids prevention epidemic cdc information gov health infection |
| Health (medical care) | hospital patient doctor medical care news information health |
| PS (elections) | party democratic political communist socialist republican labor news |
| PS (army) | force navy naval air mil commander news fleet web reserve |
| Sports (swimming) | swimming swim swimmers help information coaching technique |
| Sports (football) | football fans game nba playoff story players assault adidas university |

* Key: CAE = Culture-Arts-Entertainment, ES = Education-Science, PS = Politics-Society

Since GSDMM found significantly more topics, it was able to uncover finer-grained topics. Thus, in such cases where a brief overview is desired, the model producing the smaller number of topics might be preferable. Where more detail is desired, one can opt for a model that produces more topics.

## Conclusions and Future Work

Despite the lack of attention on the Poisson distribution in topic modelling, we have shown its utility in modelling short text. We proposed a new topic model for short text, the Gamma-Poisson mixture (GPM) topic model and performed extensive experimentation in order to investigate its properties empirically. In addition, we also derived a collapsed Gibbs sampler for the model.

As is well-known in the field topic modelling, the selection of the appropriate number of topics is a challenge. Our new topic model is able to address this in its ability to automatically select the number of topics. This is achieved via the use of the collapsed Gibbs sampler. We also showed that our model was able to find estimates that were close to the true number of topics on labelled corpora. A further benefit of the collapsed Gibbs sampler, is that it also converges very quickly, thus evading the need for long burn-in periods as is typical in the application of traditional Gibbs samplers.

In addition, GPM possesses the flexibility of allowing the user to adjust the number of topics found as required. It also tends to produce consistent results with little variation. Furthermore, when compared with the GSDMM, the GPM outperformed the GSDMM. Firstly, the number of topics found by GPM was closer to the true value. Secondly, GPM was able to find topics with higher average coherence scores, thus making it a good option for topic modelling on short text.

There are many avenues for future work related to the GPM. We plan to assess the GPM on other performance measures, such as classification accuracy in end-to-end classification. We will also perform further experimentation to compare it against other short text topic models.

## Appendix

### The derivation of collapsed sampler for Gamma-Poisson mixture model

A summary of the notation that will be used is given in Table 1.

Since the topic estimates are only dependent on the topic assignments, it is only necessary to sample the topic assignment for each document. This is achieved by sampling from the conditional probability of a document belonging to a class,

$$p\big(z_m = z \big| \mathbf{z}^{(m)}, \mathbf{x}, \boldsymbol{\alpha}, \boldsymbol{\beta}, \boldsymbol{\gamma}\big) = \frac{p(\mathbf{x}, \mathbf{z} | \boldsymbol{\alpha}, \boldsymbol{\beta}, \boldsymbol{\gamma})}{p\big(\mathbf{x}, \mathbf{z}^{(m)} \big| \boldsymbol{\alpha}, \boldsymbol{\beta}, \boldsymbol{\gamma}\big)} \propto \frac{p(\mathbf{x}, \mathbf{z} | \boldsymbol{\alpha}, \boldsymbol{\beta}, \boldsymbol{\gamma})}{p\big(\mathbf{x}^{(m)}, \mathbf{z}^{(m)} \big| \boldsymbol{\alpha}, \boldsymbol{\beta}, \boldsymbol{\gamma}\big)}, \tag{A1}$$

where the superscript $(m)$ is used to indicate that document $\mathbf{x}_m$ is excluded from $\mathbf{x}$ and $\mathbf{z}$. $\boldsymbol{\alpha} = [\alpha_1, \alpha_2, \dots, \alpha_V]$ and $\boldsymbol{\beta} = [\beta_1, \beta_2, \dots, \beta_V]$ are the hyperparameters of the Gamma prior, whereas $\boldsymbol{\gamma} = [\gamma_1, \gamma_2, \dots, \gamma_K]$ denotes the hyperparameters of the Dirichlet prior. In order to define Equation A1, we need to find $p(\mathbf{x}, \mathbf{z} | \boldsymbol{\alpha}, \boldsymbol{\beta}, \boldsymbol{\gamma})$.

Owing to conditional independence between $\mathbf{x}$ and $\mathbf{z}$, it follows that

$$p(\mathbf{x}, \mathbf{z} | \boldsymbol{\alpha}, \boldsymbol{\beta}, \boldsymbol{\gamma}) = p(\mathbf{x} | \mathbf{z}, \boldsymbol{\alpha}, \boldsymbol{\beta}) p(\mathbf{z} | \boldsymbol{\gamma}). \tag{A2}$$

It was shown in [6] that the second term on the right-hand side of Equation A2 simplifies to

$$p(\boldsymbol{z}|\boldsymbol{\gamma}) = \frac{\Delta(\boldsymbol{m}+\boldsymbol{\gamma})}{\Delta(\boldsymbol{\gamma})}, \tag{A3}$$

where $\boldsymbol{m} = [m_1, m_2, \ldots, m_K]$ and $m_k$ denotes the number of documents assigned to the $k^{\text{th}}$ topic. Using the same $\Delta$ notation as in [6], it follows that $\Delta(\boldsymbol{\gamma}) = \frac{\prod_{k=1}^K \Gamma(\gamma_k)}{\Gamma(\sum_{k=1}^K \gamma_k)}$ and $\Delta(\boldsymbol{m}+\boldsymbol{\gamma}) = \frac{\prod_{k=1}^K \Gamma(m_k+\gamma_k)}{\Gamma(\sum_{k=1}^K (m_k+\gamma_k))}$.

Now, the first term on the right-hand side of Equation A2, can be expressed as

$$p(\boldsymbol{x}|\boldsymbol{z}, \boldsymbol{\alpha}, \boldsymbol{\beta}) = \int p(\boldsymbol{x}|\boldsymbol{z}, \boldsymbol{\lambda})p(\boldsymbol{\lambda}|\boldsymbol{\alpha}, \boldsymbol{\beta})\, d\boldsymbol{\lambda}. \tag{A4}$$

Under GPM, documents and words are assumed to be independent. In addition, the word counts are assumed to follow a Poisson distribution. Thus, given the topics, the corpus can be modelled as

$$p(\boldsymbol{x}|\boldsymbol{z}, \boldsymbol{\lambda}) = \prod_{m=1}^M \prod_{v=1}^V p(x_{mv}|\lambda_{kv}) = \prod_{m=1}^M \prod_{v=1}^V \frac{\lambda_{kv}^{x_{mv}} e^{-\lambda_{kv}}}{x_{mv}!}. \tag{A5}$$

In order to simplify further derivation of the collapsed Gibbs sampler, Equation A5 can be re-expressed by the introduction of $m_k$, the number of documents assigned to the $k^{\text{th}}$ topic, and $n_{kv}$, the number of times word $v$ is observed in topic $k$, as follows:

$$p(\boldsymbol{x}|\boldsymbol{z}, \boldsymbol{\lambda}) = \prod_{k=1}^K \prod_{v=1}^V \frac{\lambda_{kv}^{n_{kv}} e^{-m_k \lambda_{kv}}}{\boldsymbol{x}!}, \tag{A6}$$

where $\boldsymbol{x}! = \prod_{m=1}^M \prod_{v=1}^V x_{mv}!$.

Now, by assuming a Gamma distribution for $\boldsymbol{\lambda}$ and substituting Equation A6 into Equation A4, we obtain

$$
\begin{aligned}
p(\boldsymbol{x}|\boldsymbol{z}, \boldsymbol{\alpha}, \boldsymbol{\beta}) &= \int p(\boldsymbol{x}|\boldsymbol{z}, \boldsymbol{\lambda})p(\boldsymbol{\lambda}|\boldsymbol{\alpha}, \boldsymbol{\beta})\, d\boldsymbol{\lambda} \\
&= \int \prod_{k=1}^K \prod_{v=1}^V \frac{\lambda_{kv}^{n_{kv}} e^{-m_k \lambda_{kv}}}{\boldsymbol{x}!} \times \frac{\lambda_{kv}^{\alpha_v-1} e^{-\frac{\lambda_{kv}}{\beta_v}}}{\Gamma(\alpha_v)\beta_v^{\alpha_v}}\, d\lambda_{kv} \\
&= \prod_{k=1}^K \prod_{v=1}^V \frac{1}{\boldsymbol{x}!\,\Gamma(\alpha_v)\beta_v^{\alpha_v}} \int \lambda_{kv}^{n_{kv}+\alpha_v-1} e^{-\lambda_{kv}\left(m_k+\frac{1}{\beta_v}\right)}\, d\lambda_{kv} \\
&= \prod_{k=1}^K \prod_{v=1}^V \frac{1}{\boldsymbol{x}!\,\Gamma(\alpha_v)\beta_v^{\alpha_v}} \times \Gamma(n_{kv}+\alpha_v)\left(\frac{\beta_v}{m_k\beta_v+1}\right)^{n_{kv}+\alpha_v} \\
&= \prod_{k=1}^K \prod_{v=1}^V \frac{\Gamma(n_{kv}+\alpha_v)}{\boldsymbol{x}!\,\Gamma(\alpha_v)} \times \frac{\beta_v^{n_{kv}}}{(m_k\beta_v+1)^{n_{kv}+\alpha_v}}. \tag{A7}
\end{aligned}
$$

The integral is solved by multiplying the equation by a constant equal to 1. The result is a gamma distribution with parameters $n_{kv} + \alpha_v$ and $\frac{\beta_v}{m_k\beta_v + 1}$. By substituting Equation A3 and A7, Equation A2 can now be written as

$$p(\boldsymbol{x}, \boldsymbol{z} | \boldsymbol{\alpha}, \boldsymbol{\beta}, \boldsymbol{\gamma}) = p(\boldsymbol{x} | \boldsymbol{z}, \boldsymbol{\alpha}, \boldsymbol{\beta}) p(\boldsymbol{z} | \boldsymbol{\gamma})$$

$$= \frac{\Delta(\boldsymbol{m} + \boldsymbol{\gamma})}{\Delta(\boldsymbol{\gamma})} \prod_{k=1}^{K} \prod_{v=1}^{V} \frac{\Gamma(n_{kv} + \alpha_v)}{\boldsymbol{x}! \, \Gamma(\alpha_v)} \times \frac{\beta_v^{n_{kv}}}{(m_k\beta_v + 1)^{n_{kv} + \alpha_v}}. \quad (A8)$$

The derivation of the conditional distribution in Equation A1 can now be concluded by substituting Equation A8 as follows:

$$p(z_m = z | \boldsymbol{z}^{(m)}, \boldsymbol{x}, \boldsymbol{\alpha}, \boldsymbol{\beta}, \boldsymbol{\gamma})$$

$$\propto \frac{p(\boldsymbol{x}, \boldsymbol{z} | \boldsymbol{\alpha}, \boldsymbol{\beta}, \boldsymbol{\gamma})}{p(\boldsymbol{x}^{(m)}, \boldsymbol{z}^{(m)} | \boldsymbol{\alpha}, \boldsymbol{\beta}, \boldsymbol{\gamma})}$$

$$\propto \frac{\Delta(\boldsymbol{m} + \boldsymbol{\gamma})}{\Delta(\boldsymbol{m}^{(m)} + \boldsymbol{\gamma})} \times \prod_{v=1}^{V} \left( \frac{\Gamma(n_{zv} + \alpha_v)}{\Gamma(n_{zv}^{(m)} + \alpha_v)} \right) \left( \frac{\beta_v^{n_{zv}}}{\beta_v^{n_{zv}^{(m)}}} \right) \left( \frac{\boldsymbol{x}^{(m)}! \, \Gamma(\alpha_v)}{\boldsymbol{x}! \, \Gamma(\alpha_v)} \right) \left( \frac{\left(m_z^{(m)}\beta_v + 1\right)^{n_{zv}^{(m)} + \alpha_v}}{(m_z\beta_v + 1)^{n_{zv} + \alpha_v}} \right)$$

$$\propto \frac{m_z^{(m)} + \gamma_z}{M - 1 + \sum_{k=1}^{K} \gamma_k} \times \prod_{v=1}^{V} \prod_{j=1}^{x_{mv}} \left( n_{zv}^{(m)} + \alpha_v + j - 1 \right) \times \beta_v^{x_{mv}} \times \frac{1}{x_m!} \times \frac{\left(m_z^{(m)}\beta_v + 1\right)^{n_{zv}^{(m)} + \alpha_v}}{\left(m_z^{(m)}\beta_v + \beta_v + 1\right)^{n_{zv}^{(m)} + x_{mv} + \alpha_v}},$$

$$(A9)$$

where $n_{zv} = n_{zv}^{(m)} + x_{mv}$ and $m_z = m_z^{(m)} + 1$. We also make use of the fact that the $\Gamma$ function has the property that $\frac{\Gamma(x+m)}{\Gamma(x)} = \prod_{j=1}^{m}(x + j - 1)$. If it is assumed that $\alpha_v = \alpha$, $\beta_v = \beta$ and $\gamma_k = \gamma$ for all $v$ and $k$, then Equation A9 simplifies to

$$p(z_m = z | \boldsymbol{z}^{(m)}, \boldsymbol{x})$$

$$\propto \frac{m_z^{(m)} + \gamma_z}{M - 1 + K\gamma} \times \frac{\beta^{N_m}}{x_m!} \times \frac{\left(m_z^{(m)}\beta + 1\right)^{n_z^{(m)} + V\alpha}}{\left(m_z^{(m)}\beta + \beta + 1\right)^{n_z^{(m)} + N_m + V\alpha}} \times \prod_{v=1}^{V} \prod_{j=1}^{x_{mv}} \left( n_{zv}^{(m)} + \alpha + j - 1 \right), \quad (A10)$$

where $N_m$ is the length of $m^{\text{th}}$ document and $\sum_{v=1}^{V} n_{zv} = n_z$ denotes the total number of documents in topic $k$.

Lastly, after sampling from Equation A9 or A10 until convergence, the lambda parameters, which give the topic distributions, are estimated by the posterior means. The posterior is given by

$$p(\lambda_{kv} | \boldsymbol{x}, \boldsymbol{z}, \boldsymbol{\alpha}, \boldsymbol{\beta}) \propto \lambda_{kv}^{n_{kv} + \alpha_v - 1} e^{-\lambda_{kv}\left(m_k + \frac{1}{\beta_v}\right)}.$$

It follows that $\lambda_{kv} \sim GAM\left(n_{kv} + \alpha_v, \frac{\beta_v}{m_k \beta_v + 1}\right)$ and the topic distribution estimates are given by

$$\hat{\lambda}_{kv} = \frac{n_{kv} + \alpha_v}{\left(m_k + \frac{1}{\beta_v}\right)}.$$

The top words that describe topic $k$ are the words with the highest expected frequencies, $\hat{\lambda}_{kv}$.

## Data Availability

All datasets can be obtained from https://github.com/qiang2100/STTM.

## Conflict of Interest

The authors declare that there are no conflicts of interest regarding the publication of this paper.

## Funding Statement

This work was performed as part of the employment of the authors by the University of Pretoria and was supported by the Centre for Artificial Intelligence Research (CAIR).